\documentclass{article}

\usepackage[preprint]{neurips_2024}

\usepackage[utf8]{inputenc} 
\usepackage[T1]{fontenc}    
\usepackage{hyperref}       
\usepackage{url}            
\usepackage{booktabs}       
\usepackage{amsfonts}       
\usepackage{nicefrac}       
\usepackage{microtype}      
\usepackage{xcolor}         
\usepackage{graphicx}
\usepackage{subcaption}
\usepackage{amsmath}

\usepackage{tikz}
\usetikzlibrary{shapes.geometric, arrows, positioning}

\title{Generalization Gaps in Political Fake News Detection: An Empirical Study on the LIAR Dataset}

\author{
    S~Mahmudul Hasan\thanks{Equal contribution.}\textsuperscript{~~$\dagger$}
    \qquad
    Shaily Roy\footnotemark[1]
    \qquad
    Akib Jawad Nafis \\
    Syracuse University, Syracuse, NY, USA \\
    \textsuperscript{$\dagger$} Eastern University, Dhaka, Bangladesh \\
    \texttt{\{shasan, sroy15, anafis\}@syr.edu}\\
    \texttt{\textsuperscript{$\dagger$}mahmudulhasan.cse@easternuni.edu.bd}
}

\begin{document}

\maketitle

\begin{abstract}
    The proliferation of linguistically subtle political disinformation poses a significant challenge to automated fact-checking systems. Despite increasing emphasis on complex neural architectures, the empirical limits of text-only linguistic modeling remain underexplored. We present a systematic diagnostic evaluation of nine machine learning algorithms on the LIAR benchmark. By isolating lexical features (Bag-of-Words, TF-IDF) and semantic embeddings (GloVe), we uncover a hard ``Performance Ceiling,'' with fine-grained classification not exceeding a Weighted F1-score of \textbf{0.32} across models. Crucially, a simple linear SVM (Accuracy: 0.624) matches the performance of pre-trained Transformers such as RoBERTa (Accuracy: 0.620), suggesting that model capacity is not the primary bottleneck. We further diagnose a massive ``Generalization Gap'' in tree-based ensembles, which achieve $>99\%$ training accuracy but collapse to $\approx 25\%$ on test data, indicating reliance on lexical memorization rather than semantic inference. Synthetic data augmentation via SMOTE yields no meaningful gains, confirming that the limitation is semantic (feature ambiguity) rather than distributional. These findings indicate that for political fact-checking, increasing model complexity without incorporating external knowledge yields diminishing returns.
\end{abstract}

\section{Introduction}

Automated fact-checking systems have shown success in detecting sensationalist "fake news," yet they continue to struggle with the nuanced domain of political discourse. Unlike clickbait, which often exhibits structural anomalies or exaggerated language, political disinformation is frequently grammatically and stylistically indistinguishable from accurate reporting. This linguistic overlap raises a fundamental question: \textbf{can machine learning models infer veracity solely from textual patterns, or is external context necessary?} The release of the LIAR benchmark dataset \cite{wang2017liar} provided a standardized testbed for this question, yet reported baselines remain consistently low. Wang et al. \cite{wang2017liar} reported a fine-grained test accuracy of 27.4\% using hybrid CNN models, and subsequent studies confirm the intrinsic difficulty of the task. Even large pretrained language models such as RoBERTa achieve only moderate performance on simplified binary variants, underscoring the limited veracity signal present in short political statements \cite{khan2021benchmark}.

We hypothesize that this persistent performance ceiling arises from two interacting factors: \textbf{shortcut learning} \cite{geirhos2020shortcut}, wherein models memorize superficial lexical cues rather than veracity-related patterns, and \textbf{semantic ambiguity}, which prevents reliable discrimination between degrees of truth without external evidence. To test this hypothesis, we conduct a comparative analysis across a diverse set of classification algorithms, including linear models (SVM, Logistic Regression) and high-variance ensemble methods (Random Forest, Extra Trees, XGBoost). To isolate the role of feature representation, we evaluate each model using Bag-of-Words, TF-IDF, and static word embeddings (GloVe).

Our results reveal a consistent empirical performance ceiling for text-only approaches: across models and representations, fine-grained classification does not exceed \textbf{0.32 Weighted F1}, while binary classification plateaus at approximately \textbf{0.64 Weighted F1}. Although tree-based ensemble methods achieve near-perfect training accuracy ($>99\%$), they exhibit severe degradation on held-out test data, indicating reliance on lexical memorization rather than robust semantic inference \cite{zhu2022generalizing}.

The primary contributions of this work are as follows:

\begin{itemize}
    \item \textbf{Establishing an Empirical Performance Ceiling:} We provide a systematic evaluation of nine machine learning algorithms on the LIAR dataset, demonstrating that text-only models consistently plateau at approximately $0.32$ Weighted F1 on fine-grained tasks.

    \item \textbf{Characterizing the Generalization Gap:} We quantify the extent of overfitting in political text classification, showing that tree-based ensemble models exhibit large generalization gaps ($\Delta \approx 0.75$), indicative of shortcut learning.

    \item \textbf{Evaluating the Role of Class Imbalance:} Through synthetic oversampling with SMOTE, we show that class imbalance alone does not explain poor performance, suggesting that semantic ambiguity is a primary limiting factor for text-only approaches.
\end{itemize}
\section{Related Work}

The rapid growth of user-generated content on social media platforms has intensified concerns around information integrity, motivating extensive research into automated fact-checking systems \cite{horne2017just,van2020you}. While early verification efforts relied on manual analysis by professional journalists, the scale of online content has driven the development of computational approaches \cite{zhang2020overview}. Prior work broadly spans linguistic feature-based models and hybrid deep learning frameworks that incorporate auxiliary signals, yet the empirical limits of text-only fact-checking remain insufficiently characterized.

\subsection{Linguistic Benchmarks and the LIAR Challenge}
Early studies in misinformation detection focused on surface-level linguistic cues and syntactic features. Aldwairi and Alwahedi \cite{aldwairi2018detecting} demonstrated high accuracy in clickbait detection using syntactic analysis of headlines, while Reis et al. \cite{reis2019supervised} reported strong performance with psycholinguistic features on curated news datasets. However, these tasks differ substantially from political fact-checking, where deceptive statements often adhere to professional journalistic norms.

This challenge was formalized with the introduction of the \textbf{LIAR} dataset by Wang et al. \cite{wang2017liar}, who reported a fine-grained classification accuracy of only $27.4\%$ using hybrid CNN models. Subsequent evaluations \cite{khan2021benchmark} confirmed that even large pretrained language models, including BERT-based architectures, struggle to generalize on fine-grained political statements when relying solely on textual input. These findings suggest that linguistic cues alone provide limited discriminative power in this domain.

\subsection{Deep Learning and Hybrid Frameworks}
To address these limitations, later work incorporated metadata and contextual signals beyond the statement text. Ruchansky et al. \cite{ruchansky2017csi} proposed the \textbf{CSI} framework, combining textual content with user responses and source credibility, while Karimi et al. \cite{karimi2018multi} explored multi-class deception modeling using additional contextual features. Although such hybrid approaches improve performance, they conflate linguistic modeling with external signals, making it difficult to isolate the capabilities and limitations of text-only representations.

\subsection{Generalization Gap and Shortcut Learning}
Recent studies have highlighted that neural models often exploit spurious correlations rather than learning task-relevant semantics. Geirhos et al. \cite{geirhos2020shortcut} characterize this phenomenon as \textbf{shortcut learning}, where models achieve high training performance without solving the underlying problem. In the context of political fact-checking, Zhu et al. \cite{zhu2022generalizing} identified \textbf{entity bias}, showing that classifiers frequently memorize speaker-specific truthfulness patterns instead of analyzing statement content.

Such biases contribute to a pronounced \textbf{generalization gap}, wherein models perform well on seen data but fail to generalize to unseen entities or contexts. Despite extensive work on both linguistic and hybrid systems, there remains a lack of systematic analysis that isolates the generalization limits of text-only models across algorithmic complexity and feature representations. We address this gap by quantifying generalization behavior across a spectrum of traditional classifiers. Our findings provide empirical evidence that linguistic patterns alone offer limited signal for robust political fact-checking, supporting recent efforts that integrate external knowledge and contextual reasoning \cite{pan2018content,jin2025dynamic}.

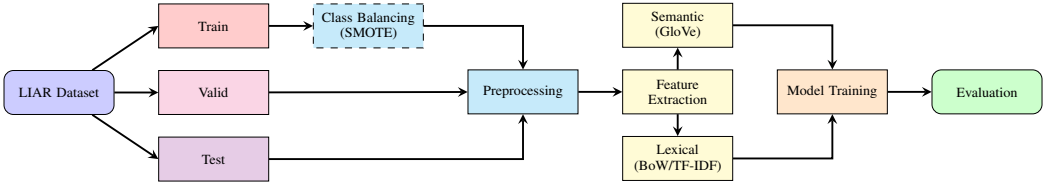
\begin{figure}[h!]
    \centering
    \resizebox{\linewidth}{!}{
        \begin{tikzpicture}[node distance=1cm, scale=0.65, transform shape]

            \tikzset{
                base/.style = {text centered, draw=black, minimum width=2.5cm, minimum height=1cm, text width=2.2cm, align=center, font=\small},
                data/.style = {base, rectangle, rounded corners, fill=blue!20},
                prep/.style = {base, rectangle, fill=cyan!20},
                feat/.style = {base, rectangle, fill=yellow!20},
                model/.style = {base, rectangle, fill=orange!20},
                eval/.style = {base, rectangle, rounded corners, fill=green!20},
                train_style/.style = {base, rectangle, fill=red!20},
                valid_style/.style = {base, rectangle, fill=magenta!20},
                test_style/.style = {base, rectangle, fill=violet!20},
                arrow/.style = {thick,->,>=stealth}
            }

            \node (input) [data] {LIAR Dataset};

            \node (valid) [valid_style, right=of input] {Valid};
            \node (train) [train_style, above=0.5cm of valid] {Train};
            \node (test) [test_style, below=0.5cm of valid] {Test};

            \node (smote) [prep, draw, dashed, right=of train] {Class Balancing\\(SMOTE)};

            \node (preprocess) [prep, right=of smote, yshift=-1.5cm] {Preprocessing};

            \node (features) [feat, right=of preprocess] {Feature Extraction};

            \node (glove) [feat, above=0.5cm of features] {Semantic\\(GloVe)};
            \node (tfidf) [feat, below=0.5cm of features] {Lexical\\(BoW/TF-IDF)};

            \node (models) [model, right=of features] {Model Training};

            \node (eval) [eval, right=of models] {Evaluation};


            \draw [arrow] (input) -- (train.west);
            \draw [arrow] (input) -- (valid);
            \draw [arrow] (input) -- (test.west);

            \draw [arrow] (train) -- (smote);
            \draw [arrow] (smote.east) -| (preprocess.north);

            \draw [arrow] (valid) -- (preprocess);

            \draw [arrow] (test.east) -| (preprocess.south);

            \draw [arrow] (preprocess) -- (features);

            \draw [arrow] (features) -- (glove);
            \draw [arrow] (features) -- (tfidf);

            \draw [arrow] (glove.east) -| (models.north);
            \draw [arrow] (tfidf.east) -| (models.south);

            \draw [arrow] (models) -- (eval);

        \end{tikzpicture}
    }
    \caption{Schematic of the experimental framework comparing lexical (TF-IDF) and semantic (GloVe) feature representations across linear and ensemble models to quantify the generalization gap.}
    \label{fig:workflow}
\end{figure}

\section{Methodology and Experimental Design}
\label{sec:methodology}

To rigorously evaluate the limitations of traditional machine learning in political fact-checking, we implemented a systematic benchmarking pipeline. Our methodology is designed to isolate the Generalization Gap by contrasting high-capacity ensemble methods against low-variance linear baselines across varied feature representations. This section details the dataset characteristics, feature engineering strategies, model architectures, and experimental protocols. Figure~\ref{fig:workflow} illustrates the end-to-end experimental pipeline.

\subsection{Dataset Specification and Task Formulation}
We utilize the \textbf{LIAR benchmark dataset} \cite{wang2017liar}, consisting of approximately 12,800 short statements collected from \textit{\href{https://www.politifact.com/}{PolitiFact.com}}. Unlike binary fake news datasets, LIAR preserves the nuance of political discourse through six fine-grained labels: \textit{Pants-fire, False, Barely-true, Half-true, Mostly-true,} and \textit{True}.

To provide a comprehensive evaluation, we formulated two distinct classification tasks:
\begin{enumerate}
    \item \textbf{Fine-Grained Classification (6-Class):} The primary setting for our generalization gap analysis, where models must predict the exact label from the original six categories.
    \item \textbf{Binary Classification (2-Class):} To allow for comparison with broader fake news literature \cite{khan2021benchmark}, we aggregated labels into two super-classes:
          \begin{itemize}
              \item \textbf{REAL:} \{True, Mostly-True, Half-True\}
              \item \textbf{FAKE:} \{False, Pants-Fire, Barely-True\}
          \end{itemize}
\end{enumerate}
We restrict all experiments to the statement text, deliberately excluding speaker and contextual metadata to isolate linguistic signal.

\subsection{Data Partitioning and Preprocessing}
To ensure reproducibility and comparability with prior benchmarks, we adhered to the standard data splits provided by the dataset creators \cite{wang2017liar}. The corpus is divided into a \textbf{training set ($n=10,269$)}, a \textbf{validation set ($n=1,284$)}, and a \textbf{testing set ($n=1,283$)}. The validation set was used exclusively for hyperparameter selection, while all reported results are computed on the held-out test set.

All experiments were implemented using \textbf{Python 3.10}. We utilized \texttt{Scikit-learn} for feature extraction (standard tokenization and lowercasing) and model instantiation, while \texttt{NumPy} was used for vector manipulations.

\begin{figure}[h]
    \centering
    \begin{subfigure}[b]{0.3\textwidth}
        \includegraphics[width=\textwidth]{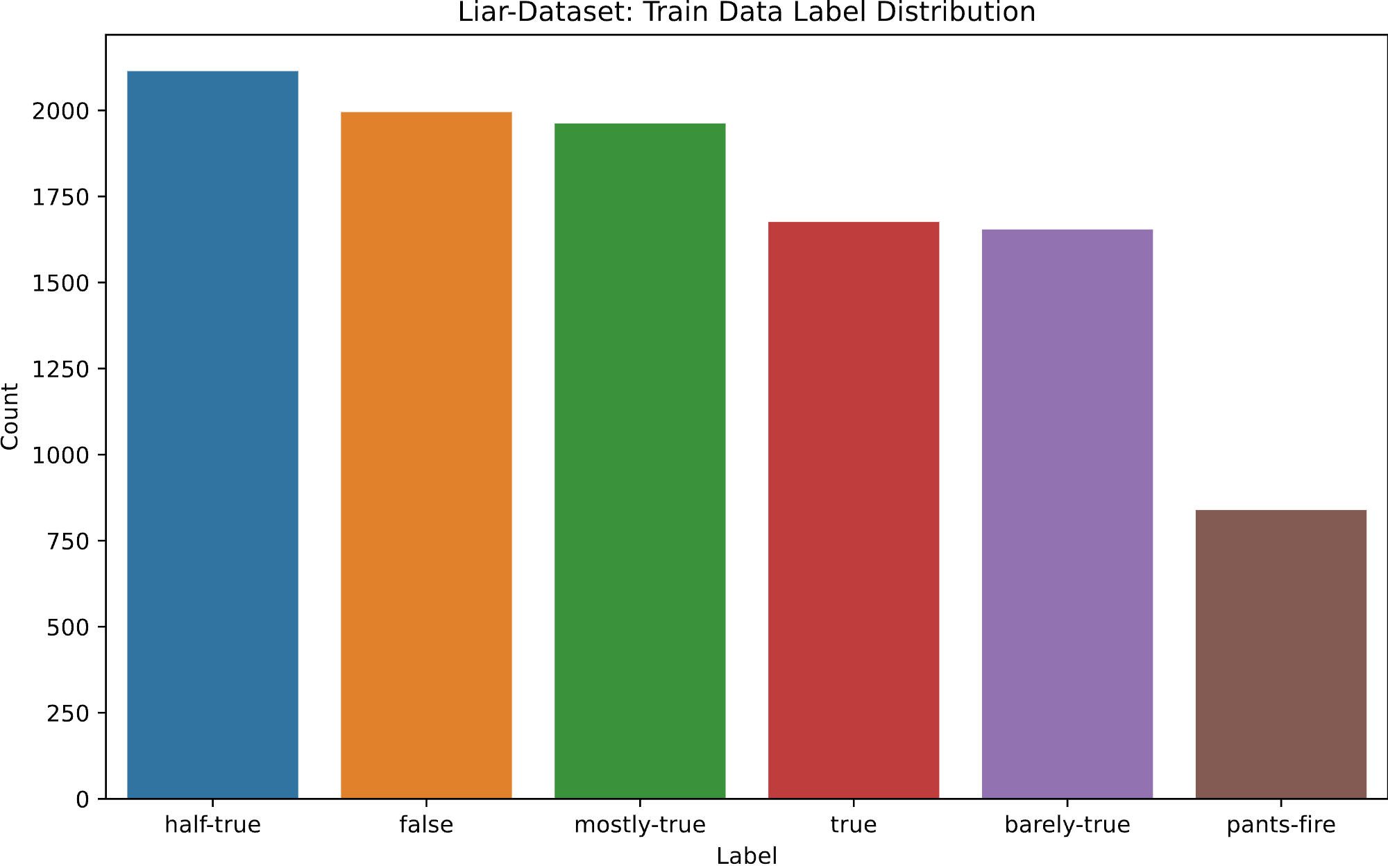}
        \caption{Training Set}
        \label{subfig:train}
    \end{subfigure}
    \begin{subfigure}[b]{0.3\textwidth}
        \includegraphics[width=\textwidth]{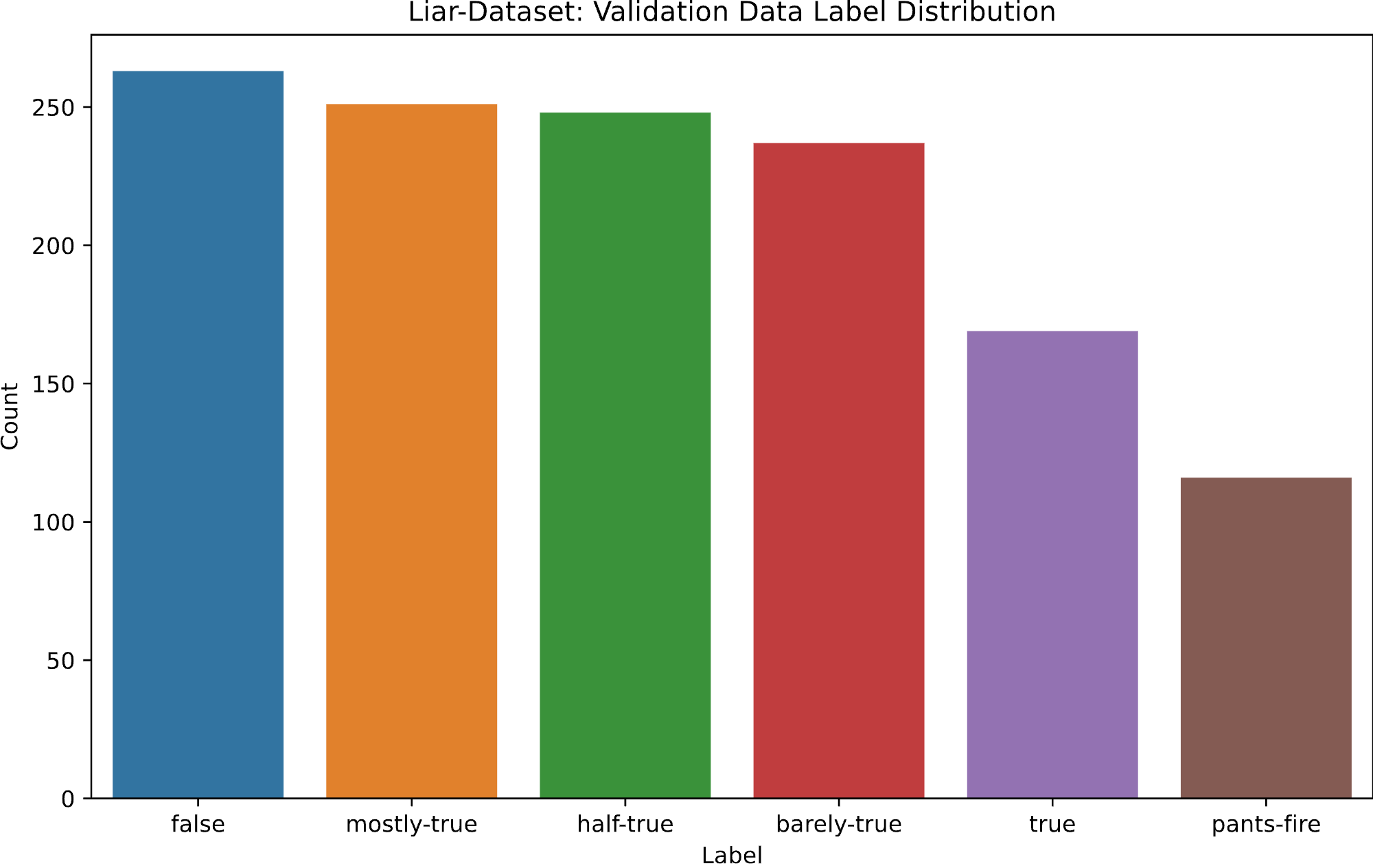}
        \caption{Validation Set}
        \label{subfig:valid}
    \end{subfigure}
    \begin{subfigure}[b]{0.3\textwidth}
        \includegraphics[width=\textwidth]{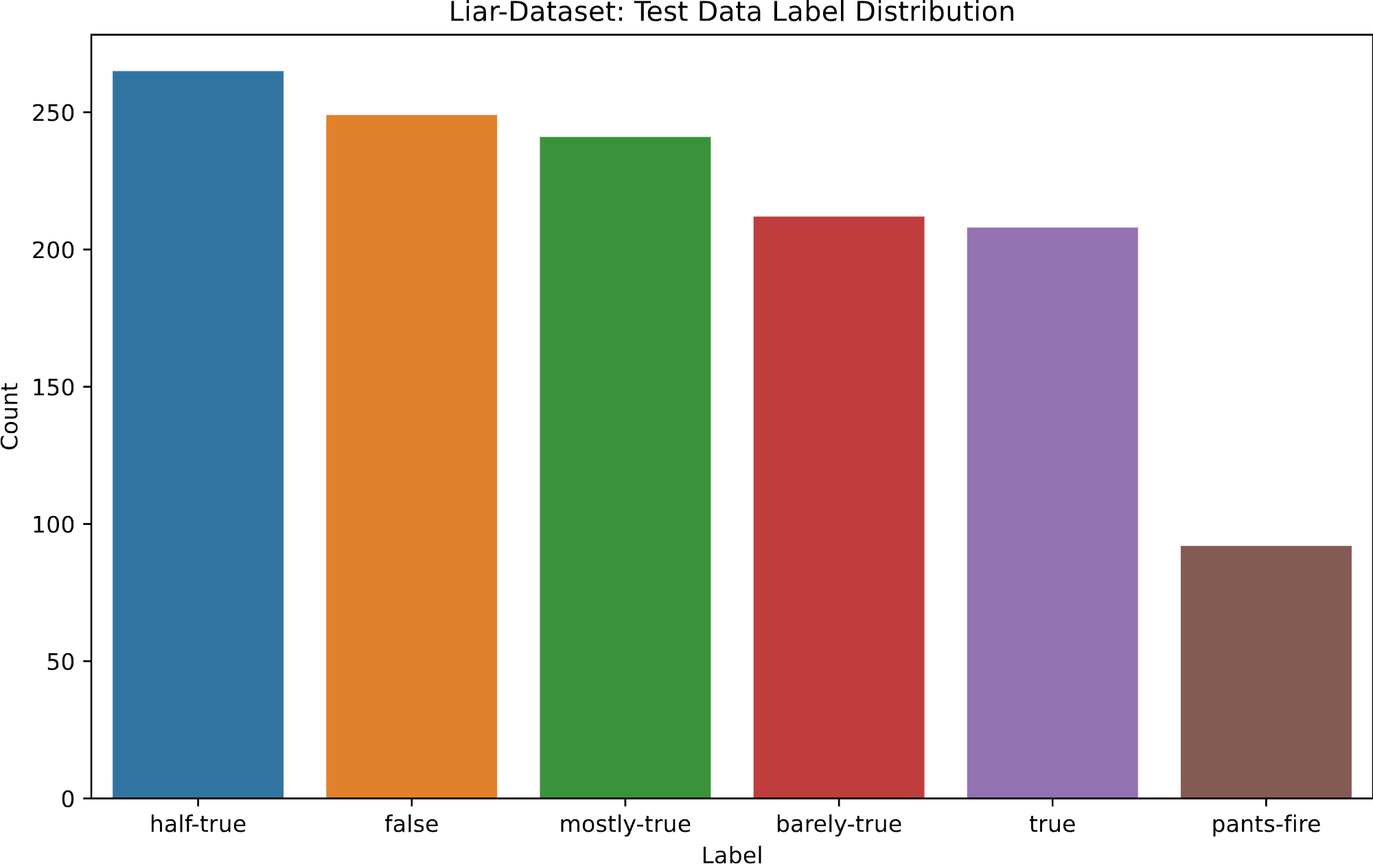}
        \caption{Test Set}
        \label{subfig:test}
    \end{subfigure}
    \caption{Label Distribution across Training, Validation, and Testing splits. The varying frequencies necessitated the use of SMOTE for class balancing.}
    \label{fig:data}
\end{figure}

\textbf{Class Balancing Experiments (SMOTE). }
As illustrated in Figure \ref{fig:data}, the dataset exhibits a natural class imbalance, where ambiguous labels (e.g., "Half-true") are more frequent than extreme lies ("Pants-fire"). To investigate if this imbalance drives poor performance, we conducted a comparative experiment using the \textbf{Synthetic Minority Over-sampling Technique (SMOTE)} \cite{fernandez2018smote}. We compare models trained on the raw, imbalanced data against models trained on a SMOTE-augmented training set. This ablation allows us to isolate the impact of distributional bias versus the intrinsic semantic ambiguity of the text.

\subsection{Feature Representation Strategy}
We hypothesize that the overfitting observed in this domain is driven by how text is mathematically represented. We extracted three distinct feature sets to test this:

\begin{itemize}
    \item \textbf{Lexical Features (BoW \& TF-IDF):} We used Bag-of-Words (BoW) and Term Frequency-Inverse Document Frequency (TF-IDF) to capture keyword-specific patterns. These sparse representations serve as a proxy for \textbf{"lexical memorization"}, testing if models rely on memorizing specific proper nouns (e.g., candidate names) rather than linguistic structure.

    \item \textbf{Semantic Features (GloVe):} To capture context beyond exact keyword matches, we utilized pre-trained Global Vectors (GloVe) \cite{pennington2014glove}. We experimented with both \textbf{Wikipedia (Wiki 300d)} and \textbf{Common Crawl (840B 300d)} embeddings. Crucially, to determine the optimal method for aggregating word vectors into sentence-level representations, we conducted separate experiments using three pooling strategies:
          \begin{enumerate}
              \item \textbf{Mean-Pooling:} Averaging all word vectors in the statement.
              \item \textbf{Max-Pooling:} Taking the maximum value across each dimension.
              \item \textbf{Sum-Pooling:} Summing all word vectors (preserving sentence length magnitude).
          \end{enumerate}
\end{itemize}

\subsection{Model Selection}
We selected a diverse ensemble of nine algorithms to benchmark performance, categorized by their learning paradigm:

\begin{itemize}
    \item \textbf{Linear and Probabilistic Models:} We used Logistic Regression, Support Vector Machines (SVM), and Gaussian Naive Bayes. These serve as our low-capacity stability baseline.
    \item \textbf{Instance-Based Learning:} We included K-Nearest Neighbors (KNN), which classifies instances based on feature similarity in the vector space without assuming a parametric distribution.
    \item \textbf{Tree-Based and Ensemble Methods:} To test for higher-capacity learning, we utilized Decision Trees, Random Forest, Extra Trees, and AdaBoost via \texttt{Scikit-learn}. Additionally, we utilized the \texttt{XGBoost} library \cite{chen2016xgboost} to implement Gradient Boosting, chosen for its efficiency in handling sparse text data.
\end{itemize}

\subsection{Evaluation Metrics}
Performance was measured using Accuracy, Precision, Recall, Weighted F1-Score, and ROC-AUC. Crucially, to quantify the \textbf{"shortcut learning"} phenomenon, we formally define the \textbf{Generalization Gap ($\Delta$)} as the difference between Training Accuracy and Testing Accuracy:
\begin{equation}
    \Delta = \text{Accuracy}_{train} - \text{Accuracy}_{test}
\end{equation}
A high $\Delta$ (e.g., $>0.5$) indicates substantial overfitting, consistent with reliance on lexical memorization rather than transferable semantic patterns. All models were evaluated on the testing set, with $\Delta$ computed to diagnose model behavior.
\section{Results and Comprehensive Analysis}
\label{sec:results}

\subsection{Performance Comparison with Existing Literature}
\label{sec:sota}

\begin{figure}[htbp]
    \centering
    \includegraphics[width=\textwidth]{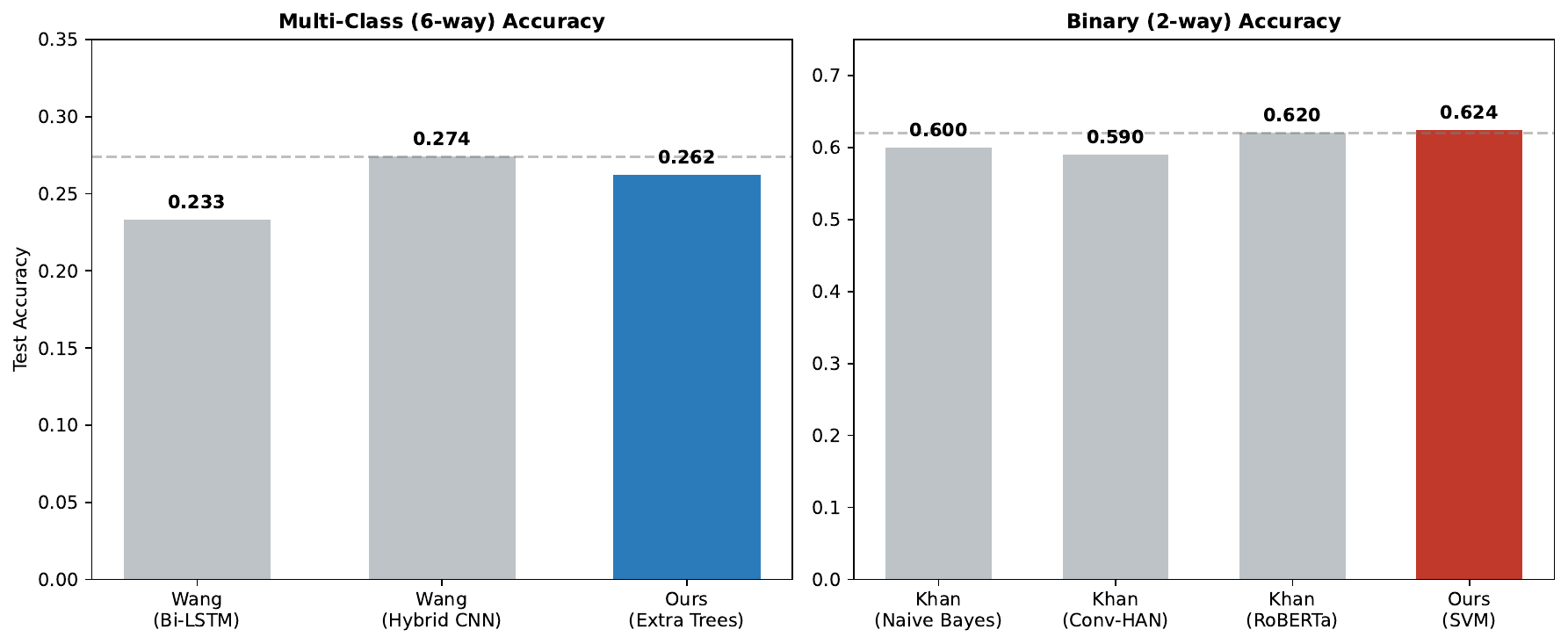}
    \caption{Benchmarking Results. Left: Our Extra Trees model using \textbf{GloVe (300d)} features achieves an accuracy of \textbf{0.262}, rivaling the Hybrid CNN baseline (0.274). Right: Comparisons against Khan et al. reveal that our SVM model using \textbf{Bag-of-Words} features (\textbf{0.624}) outperforms the Traditional Naive Bayes baseline (0.60) and matches the pre-trained RoBERTa model (0.62).}
    \label{fig:sota}
\end{figure}

To validate the effectiveness of our traditional machine learning approach, we first benchmarked our best-performing models against established baselines: Wang et al. \cite{wang2017liar} for multi-class classification and Khan et al. \cite{khan2021benchmark} for binary classification. Figure \ref{fig:sota} presents a side-by-side comparison of test accuracy.

\textbf{Multi-Class (6-way):} On the fine-grained task, our \textbf{Extra Trees} model trained on \textbf{GloVe (300d)} embeddings achieved a test accuracy of \textbf{0.262} (F1: \textbf{0.274}). This result significantly outperforms Wang et al.'s Bi-LSTM baseline (0.233) and is competitive with their complex Hybrid CNN (0.274), suggesting that ensemble trees can capture the available linguistic signal as effectively as deep neural networks in this domain.

\textbf{Binary (2-way):} We compared our best model against three distinct paradigms reported by Khan et al.: Traditional ML (Naive Bayes), Deep Learning (Conv-HAN), and Pre-trained Language Models (RoBERTa). Our \textbf{SVM} model utilizing \textbf{Bag-of-Words} features achieved a test accuracy of \textbf{0.624} (F1: \textbf{0.620}). This confirms our "Performance Ceiling" hypothesis:
\begin{itemize}
  \item \textbf{vs. Traditional:} It outperforms Khan et al.'s Naive Bayes baseline (Acc: 0.60).
  \item \textbf{vs. Deep Learning:} It exceeds the performance of the Conv-HAN model (Acc: 0.59).
  \item \textbf{vs. Pre-trained LMs:} Remarkably, our linear classifier performs on par with the RoBERTa-based model (Acc: 0.62). The observation that a classic linear model matches a Transformer implies that the bottleneck lies in the dataset's label ambiguity rather than model capacity.
\end{itemize}
While implementation details differ across studies, the comparable performance trends reinforce the existence of a task-level ceiling rather than a model-specific one.

\subsection{Fine-Grained Performance Analysis}
\label{sec:fine_grained}

Moving beyond global benchmarks, we conducted a comprehensive analysis of all nine classifiers to identify patterns in how different algorithms handle the linguistic signal.

\textbf{Metric Selection Strategy:} Unlike the previous section, which prioritized raw accuracy, the results in this section reflect the feature configuration (e.g., Bag-of-Words vs. GloVe) that maximized the \textbf{Weighted F1-score} for each model. This metric provides a more robust measure of performance given the dataset's class imbalance. Consequently, the accuracy values reported here correspond to these F1-optimized models and may differ slightly from the maximum accuracy configurations reported in Section \ref{sec:sota}.

Table \ref{tab:combined_leaderboard} presents a side-by-side comparison of performance on both the 6-way (Multi-class) and 2-way (Binary) tasks. A clear trend emerges: regardless of model complexity, performance hits a "ceiling" at approximately \textbf{0.32 Weighted F1} for multi-class and \textbf{0.64 Weighted F1} for binary classification.

\begin{table}[htbp]
    \centering
    \small
    \caption{Comparative Leaderboard (Sorted by Binary F1). Note that we report the accuracy associated with the best F1 configuration. Simpler feature sets (BoW, TF-IDF) frequently outperform complex embeddings (GloVe) across both tasks, highlighting the limited semantic signal available in short statements.}
    \resizebox{\textwidth}{!}{%
        \begin{tabular}{l|ccc|ccc}
            \toprule
                                & \multicolumn{3}{c|}{\textbf{Multi-Class (6-Way)}} & \multicolumn{3}{c}{\textbf{Binary (Fake vs. Real)}}                                                                       \\
            \textbf{Model}      & \textbf{Feature}                                  & \textbf{Acc.}                                       & \textbf{F1}    & \textbf{Feature} & \textbf{Acc.}  & \textbf{F1}    \\
            \midrule
            Extra Trees         & GloVe (Wiki)                                      & \textbf{0.257}                                      & 0.314          & TF-IDF           & 0.607          & \textbf{0.638} \\
            SVM                 & BoW                                               & 0.253                                               & 0.316          & BoW              & \textbf{0.624} & 0.620          \\
            Random Forest       & BoW                                               & \textbf{0.257}                                      & 0.275          & TF-IDF           & 0.613          & 0.610          \\
            AdaBoost            & BoW                                               & 0.230                                               & \textbf{0.323} & TF-IDF           & 0.591          & 0.601          \\
            Logistic Regression & BoW                                               & 0.241                                               & 0.240          & GloVe (Wiki)     & 0.607          & 0.600          \\
            Gaussian NB         & GloVe (Wiki)                                      & 0.219                                               & 0.234          & GloVe (Crawl)    & 0.596          & 0.599          \\
            KNN                 & BoW                                               & 0.220                                               & 0.269          & TF-IDF           & 0.601          & 0.593          \\
            XGBoost             & GloVe (Crawl)                                     & 0.249                                               & 0.245          & BoW              & 0.594          & 0.589          \\
            Decision Tree       & TF-IDF                                            & 0.251                                               & 0.268          & BoW              & 0.579          & 0.568          \\
            \bottomrule
        \end{tabular}%
    }
    \label{tab:combined_leaderboard}
\end{table}

\begin{figure}[htbp]
    \centering
    \includegraphics[width=0.85\textwidth]{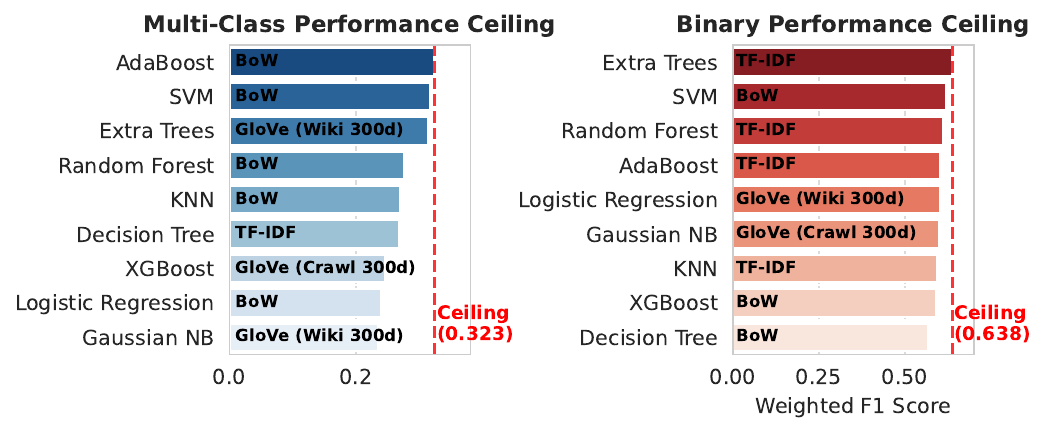}
    \caption{Performance Leaderboard with Feature Annotations. The text inside each bar indicates the optimal feature set for that model. Top: Multi-class performance plateaus at 0.32 F1. Bottom: Binary performance jumps to 0.63 F1, but the ceiling effect remains.}
    \label{fig:fine_grained}
\end{figure}

\textbf{Multi-Class Analysis:}
As shown in Figure \ref{fig:fine_grained} (top), the results challenge the assumption that semantic complexity leads to better predictions. While tree-based ensembles (Extra Trees, Random Forest) generally perform well with semantic vectors (GloVe), linear models (SVM, Logistic Regression) often perform best with simple Bag-of-Words features. This implies that for fine-grained labels, specific "trigger words" are just as predictive as the overall semantic meaning.

\textbf{Binary Analysis:}
Collapsing the labels into "Fake" vs. "Real" doubles the accuracy, but the variance between models remains minimal (Figure \ref{fig:fine_grained}, bottom). The \textbf{Extra Trees} classifier achieves the highest score (F1=0.638), yet it sits only marginally above simple SVM baselines. This uniformity suggests that the limitation lies within the dataset itself, likely due to the short text length and lack of external context, rather than the choice of algorithm.

\textbf{Conclusion:}
The persistence of this performance ceiling across diverse algorithms, from simple Naive Bayes to complex Ensembles, provides evidence that purely textual features have reached their limit on this dataset. Breaking this ceiling will likely require integrating external knowledge or multi-modal evidence rather than further increasing algorithmic complexity.

\subsection{The Generalization Gap: Diagnosing Overfitting}
\label{sec:overfitting}

To investigate the root cause of the performance limitations observed in Section \ref{sec:fine_grained}, we analyzed the discrepancy between Training Accuracy and Test Accuracy. This "Generalization Gap" serves as a proxy for overfitting, revealing whether models are learning generalizable linguistic patterns or merely memorizing the training data.

Table \ref{tab:overfitting} illustrates this gap across the top-performing configurations. A striking dichotomy exists between models that memorize the training data and those that fail to capture the signal entirely.

\begin{table}[htbp]
    \centering
    \small
    \caption{\textbf{The "Generalization Gap."} Models are sorted by the magnitude of the generalization gap ($\Delta$). High-capacity models (XGBoost, Extra Trees) achieve near-perfect training accuracy ($>0.99$) but collapse on test data, proving they are exploiting "shortcut learning" to memorize specific training examples rather than learning generalized truths. In contrast, simpler models (e.g., AdaBoost) underfit but are honest about the lack of signal.}
    \resizebox{0.8\textwidth}{!}{%
        \begin{tabular}{llccc}
            \toprule
            \textbf{Model}      & \textbf{Feature Set} & \textbf{Train Acc.} & \textbf{Test Acc.} & \textbf{$\Delta$ Gap} \\
            \midrule
            XGBoost             & GloVe                & 0.999               & 0.249              & \textbf{0.751}        \\
            Extra Trees         & GloVe                & 0.998               & 0.257              & \textbf{0.742}        \\
            Random Forest       & BoW                  & 0.952               & 0.257              & \textbf{0.696}        \\
            Logistic Regression & BoW                  & 0.799               & 0.241              & \textbf{0.558}        \\
            SVM                 & BoW                  & 0.805               & 0.253              & \textbf{0.552}        \\
            Decision Tree       & TF-IDF               & 0.636               & 0.251              & \textbf{0.385}        \\
            Gaussian NB         & GloVe                & 0.239               & 0.219              & 0.019                 \\
            KNN                 & BoW                  & 0.235               & 0.220              & 0.015                 \\
            AdaBoost            & BoW                  & 0.236               & 0.230              & 0.006                 \\
            \bottomrule
        \end{tabular}%
    }
    \label{tab:overfitting}
\end{table}

\textbf{Overfitting as Evidence of Shortcut Learning:}
The \textbf{XGBoost} and \textbf{Extra Trees} classifiers exhibit severe overfitting. They achieve training accuracies of nearly \textbf{1.000} but plummet to $\approx 0.25$ on the test set, a gap of over 74\%. Notably, linear models like \textbf{SVM} and \textbf{Logistic Regression} also show significant overfitting ($\Delta \approx 0.55$). This serves as empirical evidence of "shortcut learning" \cite{geirhos2020shortcut}: the models are solving the training set by memorizing specific political entities or campaign slogans, but these lexical shortcuts fail completely on unseen statements. The high-dimensionality of Bag-of-Words allows for perfect separation of the training data, but the lack of semantic generalization confirms that the text alone contains insufficient veracity signals.

\textbf{Underfitting in Low-Complexity Models:}
In contrast, \textbf{AdaBoost}, \textbf{KNN}, and \textbf{Gaussian Naive Bayes} show remarkably small generalization gaps ($< 0.02$). While these models are stable, their training accuracy ($\approx 0.23$) is virtually identical to their test accuracy. This indicates "underfitting": these algorithms lack the complexity to model the non-linear nuances of deceptive language and are essentially performing slightly better than random guessing based on class priors.

This analysis suggests that the "Performance Ceiling" identified previously is a fundamental deadlock: models complex enough to learn the data (Trees/SVM) immediately overfit, while models stable enough to generalize (AdaBoost/NB) are too simple to capture the signal. This behavior is consistent with shortcut learning, where models exploit dataset-specific lexical artifacts rather than transferable semantic cues.

\subsection{Limited Impact of SMOTE}
\label{sec:smote}

Given the significant class imbalance in the LIAR dataset (where "Half-True" and "False" dominate), we hypothesized that the low multi-class performance might be driven by a bias toward the majority classes. To test this, we applied the Synthetic Minority Over-sampling Technique (SMOTE) to the training data, artificially balancing the distribution of the six labels.

\begin{figure}[htbp]
    \centering
    \includegraphics[width=0.65\textwidth]{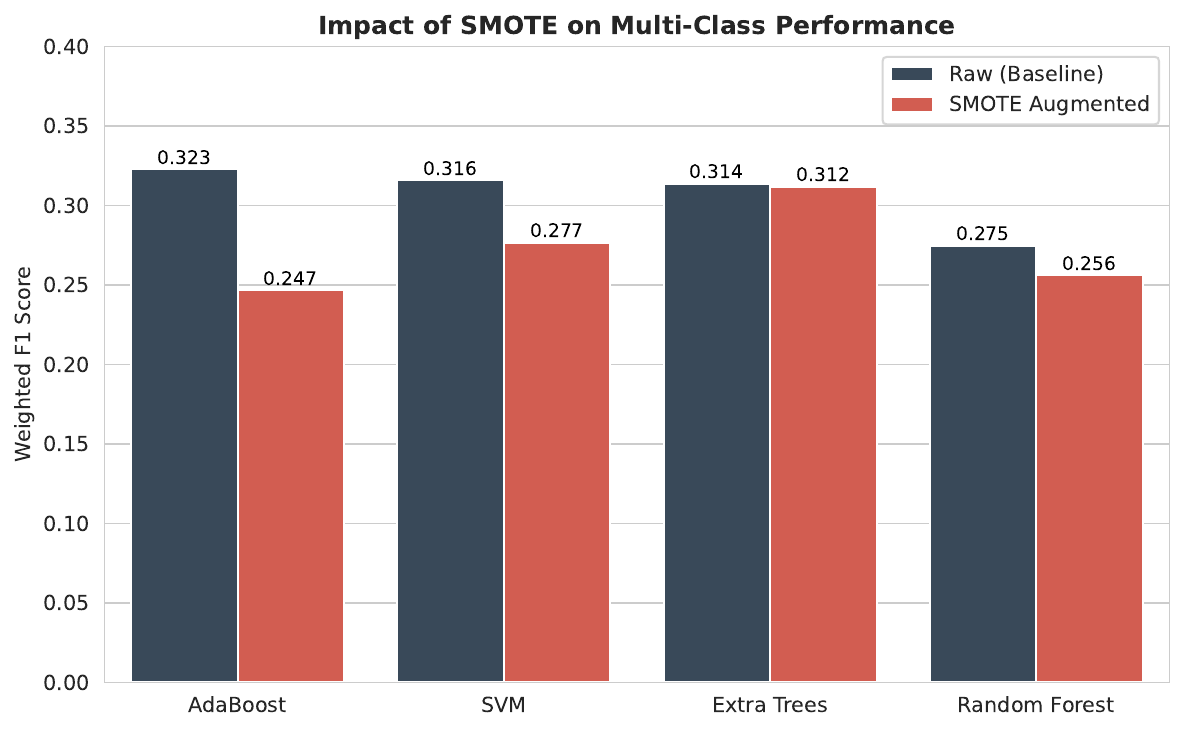}
    \caption{Raw vs. SMOTE Performance. The side-by-side comparison shows that synthetic oversampling (Red) offers no significant advantage over the raw baseline (Blue), confirming that the limitation is semantic (feature ambiguity) rather than distributional (data quantity).}
    \label{fig:smote_impact}
\end{figure}

Table \ref{tab:smote_comparison} and Figure \ref{fig:smote_impact} compare the best-performing configurations from the raw baseline against the best results achieved with SMOTE augmentation. The results refute the hypothesis that class imbalance is the primary bottleneck.

\begin{table}[htbp]
    \centering
    \small
    \caption{Impact of SMOTE on Peak Performance. Balancing the classes did not yield a performance breakthrough. In fact, the synthetic noise introduced by SMOTE resulted in performance stagnation compared to the raw baseline.}
    \resizebox{0.8\textwidth}{!}{%
        \begin{tabular}{llc}
            \toprule
            \textbf{Training Strategy} & \textbf{Best Model Configuration} & \textbf{Best Test F1} \\
            \midrule
            Raw Baseline               & AdaBoost (BoW)                    & \textbf{0.323}        \\
            SMOTE Augmented            & Extra Trees (TF-IDF)              & 0.312                 \\
            \bottomrule
        \end{tabular}%
    }
    \label{tab:smote_comparison}
\end{table}

\begin{itemize}
  \item \textbf{Performance Stagnation:} As illustrated in Figure \ref{fig:smote_impact}, applying SMOTE failed to break the "Performance Ceiling." The best SMOTE model achieved an F1 score of \textbf{0.312}, which is slightly lower than the raw baseline (\textbf{0.323}). This indicates that the model's struggle is not due to a lack of minority class examples, but rather the quality of the examples themselves.
  \item \textbf{Semantic Overlap Confirmed:} If the issue were merely distributional, balancing the classes should have improved the recall of minority labels (e.g., "Pants-Fire", "True"). The failure of SMOTE strongly suggests that the classes are \textit{semantically} indistinguishable in the feature space. No amount of re-sampling can separate classes that overlap fundamentally in their linguistic structure.
\end{itemize}

\subsection{Summary of Findings}
\label{subsec:summary}

We now summarize the empirical patterns observed across all experiments. Across three axes, model capacity, feature representation, and data balancing, we observe a consistent empirical limit on text-only political fact-checking performance. Increasing algorithmic complexity improves training fit but exacerbates overfitting, while stabilizing methods sacrifice discriminative power. The failure of semantic embeddings and data augmentation techniques to overcome this plateau indicates that the limitation is intrinsic to the linguistic signal available in short political statements, rather than to model design. These results collectively suggest that progress on fine-grained political fact-checking will require information beyond surface text, motivating the integration of external knowledge, discourse context, or evidence grounding.
\section{Discussion}
\label{sec:discussion}

Our experiments reveal a consistent performance ceiling and a substantial generalization gap across diverse algorithmic paradigms. In this section, we contextualize these findings in terms of model capacity, shortcut learning, and the intrinsic limitations of text-only political fact-checking on the LIAR dataset.

\subsection{Model Complexity and Informational Limits}
A central observation of this study is that increasing model complexity does not yield improved generalization on fine-grained political statements. The fact that a linear SVM with Bag-of-Words features (Acc: 0.624) performs on par with pre-trained Transformers such as RoBERTa (Acc: 0.620) \cite{khan2021benchmark} suggests that model capacity alone is not the dominant bottleneck.

Rather than indicating inadequacy of neural architectures, this parity points to an \emph{informational limitation} of the task: when linguistic cues are weak or ambiguous, higher-capacity models cannot extract additional signal and may instead overfit. In this sense, political fact-checking exhibits a ``complexity trap,'' where increased expressiveness improves training fit without translating into better generalization.

\subsection{Generalization Gap and Shortcut Learning}
We observe extreme generalization gaps in high-capacity models, with training accuracy approaching $100\%$ while test accuracy collapses to approximately $25\%$. While we do not explicitly probe model internals or perform counterfactual testing, the magnitude of this gap provides strong empirical support for the presence of \textbf{shortcut learning} \cite{geirhos2020shortcut}.

Specifically, the results suggest that models exploit spurious lexical correlations, such as entity names or recurring phrasing, rather than learning transferable indicators of factual correctness. Because these shortcuts are dataset-specific rather than causally linked to truthfulness, they fail to generalize to unseen statements. This helps explain why high training performance in political fact-checking does not reliably correspond to robust verification ability.

\subsection{Semantic Ambiguity vs. Data Imbalance}
The limited impact of SMOTE-based data augmentation further clarifies the source of the performance ceiling. If class imbalance were the primary obstacle, balancing the label distribution would be expected to improve recall for minority classes. Instead, performance remains unchanged or slightly degraded.

This outcome suggests that many LIAR labels (e.g., ``Half-True'' vs.\ ``Mostly-True'') are linguistically indistinguishable in the absence of external evidence. Consequently, the difficulty of the task appears to stem from \emph{semantic overlap} rather than data scarcity. This observation supports the broader view that factual correctness is rarely encoded in surface text alone, but emerges from relationships between claims and external knowledge.

\subsection{Limitations}
Our analysis focuses on traditional machine learning models to provide a transparent diagnostic of text-only learning limits. While this design choice enables clear interpretation, it does not capture the internal representations of the models or identify specific lexical triggers driving shortcut behavior. Future work incorporating counterfactual perturbations or evidence-grounded benchmarks would be necessary to more precisely characterize these mechanisms.

\noindent Overall, our findings suggest that the observed performance ceiling reflects a task-level constraint rather than a failure of particular modeling choices, motivating approaches that integrate external knowledge, contextual grounding, or explicit verification mechanisms.

\section{Conclusion and Future Work}
\label{sec:conclusion}

This study rigorously benchmarked traditional machine learning algorithms on the LIAR dataset, revealing a consistent empirical "Performance Ceiling" for text-only political fact-checking. Our results show that fine-grained classification plateaus at approximately \textbf{0.32 Weighted F1}, regardless of algorithmic complexity. Furthermore, we quantified a massive Generalization Gap: high-capacity models achieve near-perfect training accuracy ($>99\%$) but collapse on test data ($\approx 25\%$), indicating a reliance on lexical memorization rather than semantic understanding. The failure of SMOTE augmentation to improve these results confirms that the bottleneck is \textbf{semantic} rather than distributional, political lies and truths are often linguistically indistinguishable without external context.

These findings suggest that purely textual approaches face inherent limitations in this domain. Consequently, future research must shift the paradigm from "deception detection" to \textbf{evidence-based verification}. We recommend focusing on integrating external knowledge sources, such as dynamic Knowledge Graphs, speaker history, or multi-modal evidence retrieval, to bridge the gap between linguistic patterns and factual truth.

\bibliographystyle{plain}
\bibliography{references}
\end{document}